\pdfoutput=1
\documentclass[12pt,a4paper]{article}
\usepackage[T1]{fontenc}
\usepackage{lmodern}
\usepackage[english]{babel}
\usepackage{graphicx}
\usepackage{caption}
\usepackage{amsmath}
\usepackage[toc]{appendix}
\usepackage{amsfonts}
\usepackage{natbib}
\usepackage{xcolor} 
\usepackage{CJKutf8}
\usepackage{booktabs}
\usepackage{colortbl}
\usepackage{array}
\usepackage{xcolor}
\usepackage{graphicx} 
\usepackage{makecell}
\usepackage{caption}
\usepackage[colorlinks, citecolor=blue]{hyperref}    
\usepackage{authblk}

\title{\large{Comprehensive Reassessment of Large-Scale Evaluation Outcomes in LLMs:\\
	A Multifaceted Statistical Approach}}
\date{\vspace{-5ex}}

\author[1]{\footnotesize Kun Sun\thanks{\texttt{kun.sun@uni-tuebingen.de}}}
\author[2]{\footnotesize Rong Wang}
\author[3]{\footnotesize Anders Søgaard}

\affil[1]{\footnotesize Department of Linguistics, University of Tübingen, Germany}
\affil[2]{\footnotesize Institute of Natural Language Processing, Stuttgart University, Stuttgart, Germany}
\affil[3]{\footnotesize Department of Computer Science, University of Copenhagen, Copenhagen, Denmark}

\begin{document}

\maketitle

\begin{abstract}
Amidst the rapid evolution of LLMs,  the significance of evaluation in comprehending and propelling these models forward is increasingly paramount. Evaluations have revealed that factors such as scaling, training types, architectures and other factors profoundly impact the performance of LLMs. However, the extent and nature of these impacts continue to be subjects of debate because most assessments have been restricted to a limited number of models and data points. Clarifying the effects of these factors on performance scores can be more effectively achieved through a statistical lens. Our study embarks on a thorough re-examination of these LLMs, targeting the inadequacies in current evaluation methods. With the advent of a uniform evaluation framework, our research leverages an expansive dataset of evaluation results, introducing a comprehensive statistical methodology. This includes the application of ANOVA, Tukey HSD tests, GAMM, and clustering technique, offering a robust and transparent approach to deciphering LLM performance data. Contrary to prevailing findings, our results challenge assumptions about emergent abilities and the influence of given training types and architectures in LLMs. These findings furnish new perspectives on the characteristics, intrinsic nature, and developmental trajectories of LLMs. By providing straightforward and reliable methods to scrutinize and reassess LLM performance data, this study contributes a nuanced perspective on LLM efficiency and potentials. 
\end{abstract}

\small{{\bf Keywords:}{LLM evaluations, factor effects, statistical tests, emergent abilities}

\clearpage

\section{Introduction}
\label{introduction}

The advent of Large Language Models (LLMs) marks a significant milestone in the evolution of artificial intelligence. These models are revolutionizing the way we interact with technology, offering unprecedented capabilities, and consequently reshaping the AI landscape, prompting new discussions about artificial general intelligence (AGI )(\citealp{bubeck2023sparks}; \citealp{zhao2023survey}). 
The advancement of LLMs has been remarkable, yet the foundational elements that govern their operations remain somewhat of a mystery. For instance, central to this puzzle is understanding why LLMs exhibit certain advanced abilities that their smaller counterparts do not \citep{wei2022emergent}. This phenomenon underscores the urgency for thorough research to dissect the underlying factors contributing to these advanced features. 

With the rapid emergence of numerous LLMs, it has become crucial to swiftly and effectively evaluate their performance adopting reliable and standardized approaches. The extraordinary pace at which LLMs are evolving presents challenges in fully grasping their nature, characteristics, and potentials. As mentioned above, the mystery issues on LLMs could potentially be resolved through efficient and thorough evaluations. To assess the effectiveness and superiority of LLMs, a significant number of tasks and benchmarks have been introduced, aiming at empirically evaluating and analyzing their capabilities and the factors influencing their abilities (\citealp{chia2023instructeval}; \citealp{liang2022holistic}; \citealp{zhao2023survey}; \citealp{chang2023survey}; \citealp{guo2023evaluating}).
Current evaluation datasets predominantly focus on specific abilities like language understanding, reasoning, and human alignment individually. Previous research identifies several critical measures that must be considered in the evaluation of LLMs, such as accuracy, efficiency, bias, safety etc. (\citealp{liang2022holistic}; \citealp{chang2023survey}). Accuracy is paramount, encompassing not only factual correctness but also the precision of inferences and problem-solving. Efficiency is also vital. The speed at which these models generate results can be a decisive factor in their deployment for critical scenarios. Additionally, LLMs exhibit neutrality and are devoid of social biases. 
However, current LLM evaluations tend to prioritize accuracy (\citealp{fu2023chain}; \citealp{safdari2023personality}; \citealp{choi2023llms}; \citealp{yuan2023revisiting}; \citealp{li2023api}).

Recent evaluation efforts reveal several glaring issues. For instance, ``emergent abilities'' could be observed from a number of LLMs, such as GPT, PaLM and LaMDA (\citealp{wei2022emergent}; \citealp{schaeffer2023emergent}).
Some researchers found that instruction-tuning provides a broad set of advantages compared with other types of training (fine-tune, pretrained, RL-tuned etc.) (\citealp{liang2022holistic}; \citealp{chung2022scaling}; \citealp{zhao2023survey}). \cite{zhao2023survey} also reported that the small-sized open-source models perform not well on mathematical reasoning and scaling the open-source modes can improve the performance consistently. Researchers also found that some of the inconsistencies among the relationships between model size and task performance \citep{burnell2023revealing}. These findings actually are involved the overall performance of LLMs and different abilities with training types and scaling. However, the findings drawn from these studies primarily stem from observations made using a relatively small dataset. Notably, these findings have not undergone rigorous validation with a more extensive dataset. For enhanced reliability and accuracy of the results, further validation efforts could benefit from the application of comprehensive statistical methods. The following details these potential problems and challenges. 

A primary issue is the narrow range of models typically assessed in multiple tasks — often several to 30 (\citealp{yu2023mm}; \citealp{yu2023kola}; \citealp{fu2023mme}; \citealp{jiang2023structgpt};  \citealp{huang2023trustgpt}), compared to the over 120000 models available, for instance, on \texttt{Huggingface}. The limited selection fails to capture the full spectrum of LLMs, diminishing our understanding of their diverse capabilities.  For example, the limited number of LLMs may have emergent abilities. The question is whether these few models could represent the population of LLMs. As the array of LLMs expands and the availability of massive data on LLMs evaluation results, it is crucial to consider them collectively, requiring more inclusive sampling methods to ensure the findings' representativeness and reliability. When LLM evaluations frequently depend on datasets with insufficient sample sizes (often ranging from three to 30 datapoints), this raises concerns about the validity of conclusions drawn from such limited data/models.
Further, essential characteristics of LLMs, such as emergent abilities, should be analyzed using larger datasets and considering broader factors like training types and architectures, not just parameters like scale (e.g., parameter count, FLOPs). Moreover, the interplay of different LLM capabilities has not been examined. Understanding the potential interactions among their various abilities, similar to the interplay seen in human cognitive abilities (\citealp{conway2002latent}; \citealp{buehner2006cognitive}; \citealp{socher2022relationship}), remains a largely unexplored area in LLM research. Finally, the critical aspects of evaluating LLMs revolve around understanding the impact of scaling factors, training types, and architectural designs on their performance. This evaluation process bears resemblance to the assessment of human cognitive abilities, where factors like age, education, race, and sex are analyzed for their influence on cognitive skills within a population. These complex issues in both LLMs and human cognitive evaluations can be effectively addressed and validated through meticulous statistical testing and analysis. Similar challenges in other fields have been successfully addressed through the application of multiple statistical methods. 

The rapid growth of LLMs calls for more comprehensive and reliable evaluation methods. This necessitates broadening the scope of assessments, employing rigorous statistical methods. A straightforward, reliable and efficient approach is essential to accurately assess the capabilities and limitations of LLMs. To achieve this, large-scale data on evaluation results is needed, using consistent evaluation datasets and standards across numerous LLMs. Fortunately, some researchers have begun establishing platforms for this unified data collection. Once collected, both basic and advanced statistical methods could be applied to thoroughly analyze these evaluation result data. Currently, fundamental statistical techniques, such as ANOVA or $\chi$\^2 tests, are missing in testing resulting data. These analyses are crucial for understanding whether LLM performance varies significantly across different training types, architectures, and parameter sizes. Moreover, (non-)linear regression models can be employed to examine how training parameters or types affect LLM performance, and to explore the interactions among various LLM capabilities. These different statistical methods can cross-validate each other. These multifaceted statistical analyses will create a comprehensive framework, enabling an in-depth re-evaluation of the performance result data in LLMs, shown as in Fig.\ref{eva_met}.

\begin{figure}[ht]
	\vskip 0.16in
	\begin{center}
		\centerline{\includegraphics[width=0.9\textwidth]{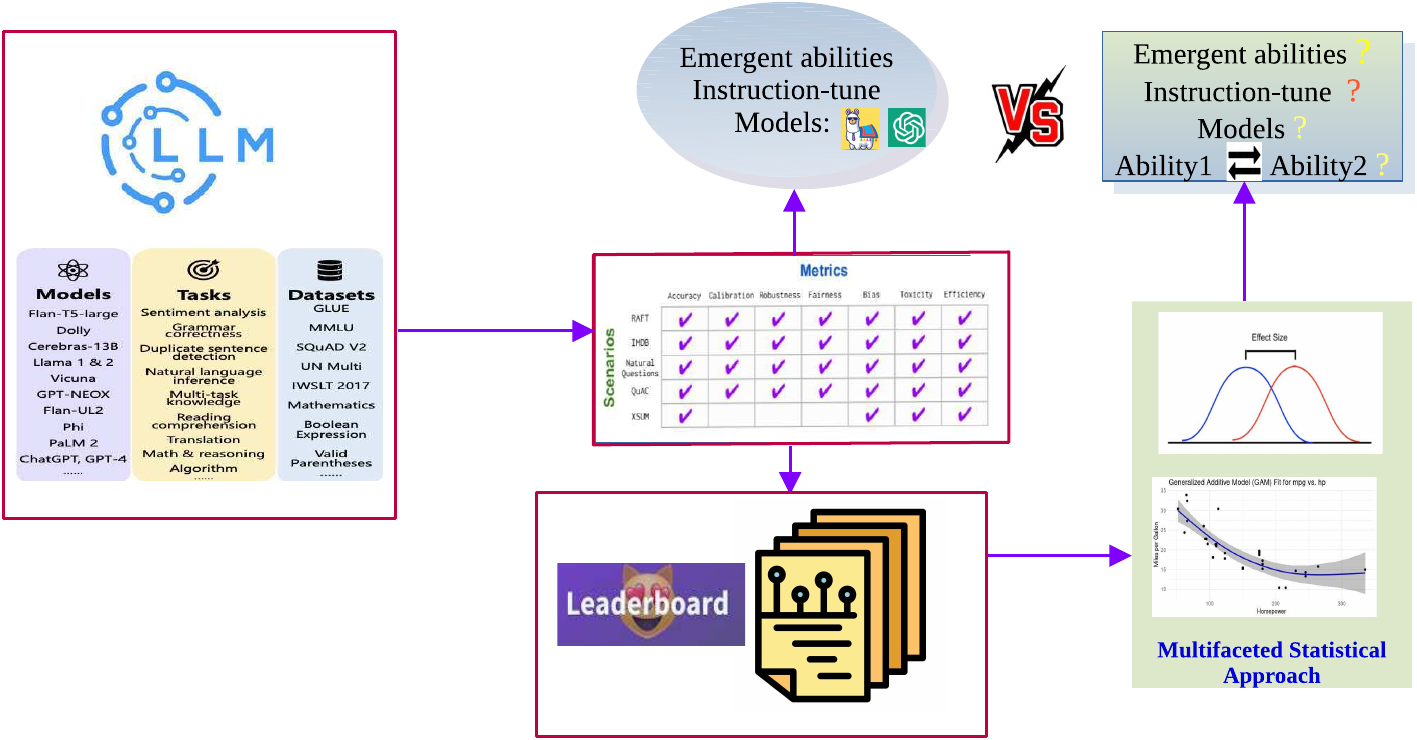}}
		\caption{Reassessment methods for massive LLMs evaluation outcomes}
		\label{eva_met}
	\end{center}
	\vskip -0.31in
\end{figure}

In short, the multifaceted statistical framework is poised to validate the core issues in evaluating LLMs, including their emergent abilities, the efficacy of training types and architectures, and the distinct advantages held by training parameter scale. This study also aims to delve into the interplay among various abilities in LLMs, investigating how they influence and interact with each other.

\section{Methods}
\label{methods}
As highlighted in the introduction, evaluations of LLMs have predominantly focused on a limited set. To address this, it is prudent to establish a centralized platform that consolidates results from various LLMs, utilizing identical datasets and evaluation criteria. Such a platform ensures that the data on LLM evaluations is comprehensive, diverse, and representative. The \texttt{Open LLM Leaderboard} serves this purpose, aiming to systematically track, rank, and assess LLMs \citep{open-llm-leaderboard} (Available at \url{huggingface.co/spaces/HuggingFaceH4/open_llm_leaderboard}).
This platform offers the extensive and cohesive dataset on LLM evaluation results. For our analysis, we extracted the data from the \texttt{Leaderboard}.

\subsection{The datasets}

The \texttt{Open LLM Leaderboard} evaluation process involves several benchmarks from \textit{the Eleuther AI Harness}\citep{eval-harness}, a unified framework for measuring the effectiveness of LLMs \citep{open-llm-leaderboard}. Table \ref{benchmark} is a brief overview of each benchmark.

\begin{table}[ht!]
	\centering
	\scalebox{0.49}{
	\begin{tabular}{|l|l|l|l|l|}
		\hline
		\textbf{Name} & \textbf{Specificity} & \textbf{Capability} & \textbf{Scale} & \textbf{Number of Shots} \\ \hline\hline
		\makecell{AI2 Reasoning Challenge (ARC)\\ \citet{clark2018think}} & Grade-School Science Questions & Knowledge & 7787 science exam questions & 25-shot \\ \hline
		\makecell{HellaSwag\\ \citet{zellers2019hellaswag}} & Commonsense Inference & \makecell{Knowledge Reasoing\\-Complex Reasoning} & 70000 questions & 10-shot \\ \hline
		\makecell{MMLU\\ \citet{hendrycks2021measuring}} & Massive Multi-Task Language Understanding & Language Comprehension & \makecell{15908 questions, \\knowledge on 57 domains} & 5-shot \\ \hline
		
		\makecell{Winogrande\\ \citet{win02019}} & Adversarial Winograd Schema Challenge & Common Sense and Reasoning & 44k questions & 5-shot \\ \hline
		\makecell{GSM8k\\ \citet{gsm8k2020}} & Grade School Math Word Problems Solving & \makecell{Symbolic Reasoning\\-Complex Reasoning} & 8.5K math questions & 5-shot \\ \hline
		\makecell{TruthfulQA\\ \citet{lin2022truthfulqa}} & Propensity to Produce Falsehoods & Human Alignment & 817 questions & 0-shot \\ \hline
	\end{tabular}}
	\caption{The benchmark evaluation datasets in the \texttt{Open LLM Leaderboard}}
	\label{benchmark}
\end{table}

\vspace{-2mm}

These renowned and extensive evaluation datasets cover a wide range of capabilities including knowledge, complex reasoning, and human alignment. Knowledge utilization is tested through datasets like ``ARC''.
Complex Reasoning is assessed via datasets like ``HellaSwag'', which emphasizes knowledge reasoning, and ``GSM8k'', centered on symbolic reasoning, with a focus on accuracy and problem-solving rates. For Human Alignment, ``TruthfulQA'' is used to gauge the truthfulness of responses from LLMs. Collectively, 
these benchmarks provide an assessment of a model's capabilities in terms of knowledge, reasoning, and some math, in various scenarios.The performance metric is recorded using scores, such as accuracy. These evaluation datasets are designed to assess multiple dimensions of cognitive ability in LLMs. 

As of \emph{January 12, 2024}, we collected the evaluation results on 1212 LLMs from the \texttt{Open LLM Leaderboard} which covered a broad spectrum of LLMs, ranging from open-source to closed-source API-accessible models. We excluded models with zero parameters, totaling unique \textbf{1186}. The dataset from the \texttt{Leaderboard} largely surpasses the size of most datasets on LLM evaluation results, which typically evaluate between a few and up to 30 models, making it a more comprehensive resource for evaluation results. This number is expected to grow as more researchers contribute their results. The data on LLMs evaluation results from the \texttt{Leaderboard} also include a diverse range of factors: \textit{architectures}, \textit{training types}, and \textit{hyperparameters}. Their (hyper)parameter counts span from \(0.01B (billion)\) to \(180B\), with an average (\(\mu\)) of \(8.19B\), a median (\(M\)) of \(6.61B\), and a standard deviation (\(\sigma\)) of \(13.73B\). The training types are categorical factors, and they are categorized into five groups with their proportions: ``fine-tune'' (63\%), ``instruction-tune''\citep{wang2023far} (15\%), ``pretrained'' (13.5\%), ``RL-tune''(Reinforcement Learning from Human Feedback tune)(1.8\%) \citep{zheng2023secrets}, and unknown (7\%). Instruction-tuned models have been further refined with additional fine-tuning using various instructions, such as task datasets, daily conversations, and synthetic instructions. There are 31 distinct architectures (categorical factors), including ``BloomForCausalLM'', ``GPT2LMHeadCustomModel'', ``LlamaForCausalLM'', and others. These 31 architectures could be further classified into 12 broader categories with their proportions: ``Bloom'' (3\%)\citep{workshop2022bloom}, ``Falcon'' (1\%)\citep{almazrouei2023falcon}, ''GLM'' (1\%)\citep{zeng2022glm}, ``GPT2'' (9\%)\citep{radford2019language}, ``GPTJ'' (3.3\%)\citep{gpt-j}, ``GPTNeo''(11\%)\citep{black2022gpt}, ``Llama'' (54\%)\citep{touvron2023llama}, ``Mistral'' (2.4\%)\citep{jiang2023mistral}, ''OPT''(3.6\%)\citep{zhang2022opt}, ``Rwkv''(1\%)\citep{peng2023rwkv}, and Other (9.9\%). 
 Additional factors, like ``precision'' and ``Hub License'', are also included in the data from the \textit{Leaderboard}. 

To ensure the robustness of our analysis, we cross-validated the other data on LLMs evaluation results, which is the supplementary dataset in the present study. This \textbf{supplementary dataset} comprises a comprehensive compilation of performance scores across several well-known LLM evaluation datasets, alongside detailed specifications of parameters and architectures. However, the supplementary dataset encompasses a relatively limited scope, covering \textbf{65} LLMs. We employed analogous re-evaluation methods to process and analyze it. The detailed outcomes of this supplementary dataset are reported in the \textbf{Appendix D}.

\subsection{Re-evaluation methods}

After gathering data from sources like the \texttt{Open LLM Leaderboard}, we applied a multi-faceted statistical approach to analyze the data. The following details the three statistical methods to re-evaluate the data on LLMs performance results across multiple dimensions.

\textbf{ANOVA and Tukey Tests}: We categorize data into groups based on factors like ``architecture'', ``training types'', and ``parameter count'', then apply ANOVA (Analysis of Variance) tests to identify significant differences across categories within each dataset. Where differences are significant, Tukey HSD (Honestly Significant Difference) tests are used for detailed pairwise comparisons. Such tests could help know whether data filtered by such factors really show significant differences on the data of performance scores. Specifically, we employed a two-step analytical method to compare differences across multiple categories within a dataset. First, we segments the data into subsets based on specific categories (such as architecture categories ``Bloom'', ``GPT2'', ``GPTJ'', etc., or training types ``fine-tuned'', ``instruction-tuned'', or parameter range scales, ``0-1.5B'', ``1.5-3B'' etc.) for various evaluation benchmark datasets, such as ``ARC'', ``HellaSwag'', and others. Each subset corresponds to a category within a given benchmark dataset. Then, we conduct ANOVA tests on these subsets to statistically evaluate the differences among the categories for each column. If the ANOVA results indicate significant differences, a post-hoc Tukey HSD test is applied for pairwise comparison between the categories. This method allows for a detailed understanding of how different categories influence the values in scores of the benchmark datasets, providing insights into the underlying patterns and relationships.

\textbf{GAMM}: Performance scores in one given evaluation dataset are treated as a \textbf{dependent variable}, while training parameters are considered an \textbf{independent variable}. Other factors, such as training types and architectures, are treated as \textbf{random variables}. Given this setup, we can use regression models to carry out comprehensive tests and analyses, exploring the relationship between the dependent variable (performance scores) and the independent variable (training parameters). Generalized Additive Mixed Models (GAMM) can primarily analyze and model complex, non-linear relationships between dependent variable and independent variables in large datasets, which is implemented by the \texttt{mgcv} package in R (\citealp{wood2017gamm}; \citealp{wood2015package}). By employing smooth terms, the method adeptly captures non-linear relationships between the dependent variables (like scores of HellaSwag, ARC) and independent variables (Parameter count). This is crucial in real-world data where relationships are rarely strictly linear. The inclusion of random effects for variables (like architectures, training types) allows for the modeling of group-specific variations. 
The method's application across different performance metrics (like HellaSwag, ARC, MMLU, etc.) demonstrates its versatility. This approach is crucial for modeling complex, non-linear relationships in our dataset, accommodating random effects and capturing variations across different groups or categories, which are often encountered in advanced data science applications.

\textbf{Clustering Analysis}: t-SNE (t-Distributed Stochastic Neighbor Embedding)\citep{van2008visualizing} effectively simplifies complex, high-dimensional data into a 2-dimensional space, making it easier to identify patterns, clusters, and relationships. By categorizing and color-coding additional variables (like range of parameters, training types etc.), it provides a deeper insight into how these variables relate to the data clusters formed in the t-SNE plot. This method is significant for its ability to transform and visualize complex datasets in a way that highlights underlying patterns and relationships. Overall, t-SNE aids in understanding how various variables (e.g., parameter, types) interact within the data clusters.  This can also provide cross-verification with the results from the above two methods.

\section{Results}
\label{results}
\subsection{Result 1: Difference analysis by parameter, and training type / architecture}



In analyzing scores from various benchmark evaluation datasets, we categorized them based on three distinct criteria: model training types, architectural frameworks, and parameter range scales. Notably, the model training types encompass five categories (fine-tuned, instruction-tuned, pretrained, RL-tuned, and unknown). The architecture models include 12 frameworks such as Bloom, Falcon, GLM, GPT2, GPTJ, GPTNeo, Llama, Mistral, OPT, Rwkv, and Others. The parameter ranges (billion) are segmented into distinct brackets, mirroring those used in the \texttt{Open LLM Leaderboard}: [0, 1.5](23.1\%), [1.5, 3](5.4\%), [3, 7](39.3\%), [7, 13](19.8\%), [13, 35](10.8\%), and [35, 80](2\%).

Our primary focus centers on the implications of parameter range scales. Employing ANOVA and Tukey's tests on scores from various benchmark datasets, we identified several significant findings (using a significance threshold defined by a \textit{p}-value less than 0.05). In an analysis of various datasets, statistical significance varies across parameter ranges. In the six evaluation datasets, many comparisons between specific parameter ranges showed insignificant differences. In the ``TruthfulQA'' dataset, certain ranges exhibited significant differences, a contrast to the other datasets. It turned out aht the range \textbf{[3,7]} consistently demonstrated significant differences across multiple datasets. This indicates that only certain parameter scales are significant in the performance of LLMs. 
The detail is seen in the \textbf{Appendix A}.


Second, We conducted a detailed examination of model training types, which include fine-tune, instruction-tune, pretrained, RL-tune, and unknown categories, utilizing ANOVA and Tukey's tests. Throughout this analysis, we consistently applied the same significance threshold for comparability and rigor. Our analysis reveals notable findings across different datasets. In the HellaSwag dataset, as well as in ARC, MMLU, TruthfulQA, and Winogrande, the differences between pretrained vs.fine-tuned and pretrained vs. instruction-tuned models are statistically significant. In the GSM8K dataset, the pretrained vs. instruction-tuned category stands out significantly. These results indicate a consistent, significant difference between pretrained and instruction-tuned models across all six evaluation datasets. Similarly, a significant distinction is observed between pretrained and fine-tuned models in these datasets. However, no significant differences are noted between instruction-tuned and fine-tuned or RL-tuned models. This suggests that while \textbf{instruction-tuned} models demonstrate benefits compared to pretrained models, their advantages are \textbf{not} as pronounced when contrasted with \textbf{fine-tuned} models. In summary, the efficacy of instruction tuning is apparent, yet it does not clearly surpass the benefits of fine-tuning.

Third, we categorized the data according to 11 architectures. Subsequently, ANOVA and Tukey's tests were conducted, applying a consistent threshold for the \textit{p}-value across all analyses. Our analysis identified significant differences between pairs of architectures, as summarized in Table \ref{arch_sign} in the \textbf{Appendix A}.  
This analysis reveals that architectures such as ``GPT2'', ``Llama'', ``Bloom'', ``Mistral'', and ``GPTNeo'' are particularly effective in executing evaluation tasks. Their recurrent appearance in significant pairings underscores their robust performance across different benchmarks.

Additionally, the correlations among the scores of the six evaluation datasets are illustrated in Fig.\ref{fig_cor_plot} in the \textbf{Appendix A}. The correlationship reveals that ``TruthfulQA'', representing human alignment, does not exhibit a high correlation with the abilities represented by the other evaluation datasets. 

\subsection{Result 2: GAMM analysis: Emergent abilities and interplay of various abilities }

The related research has identified the two defining properties of emergent abilities in LLMs: a). Sharpness, transitioning seemingly instantaneously from not present to present; b) Unpredictability, transitioning at seemingly unforeseeable model scales \citep{schaeffer2023emergent}. Actually, emergent abilities hold by default that such abilities do not decrease as the training size becomes larger. \citet{schaeffer2023emergent} utilized various metrics to scrutinize the purported emergent abilities, shown as in Fig. \ref{emg_abi},  discovering inconsistencies between these abilities and established principles. Consequently, they concluded that emergent abilities might not be an inherent aspect of scaling AI models.
\begin{figure}
	\vskip 0.16in
	\begin{center}
		\centerline{\includegraphics[width=0.9\columnwidth]{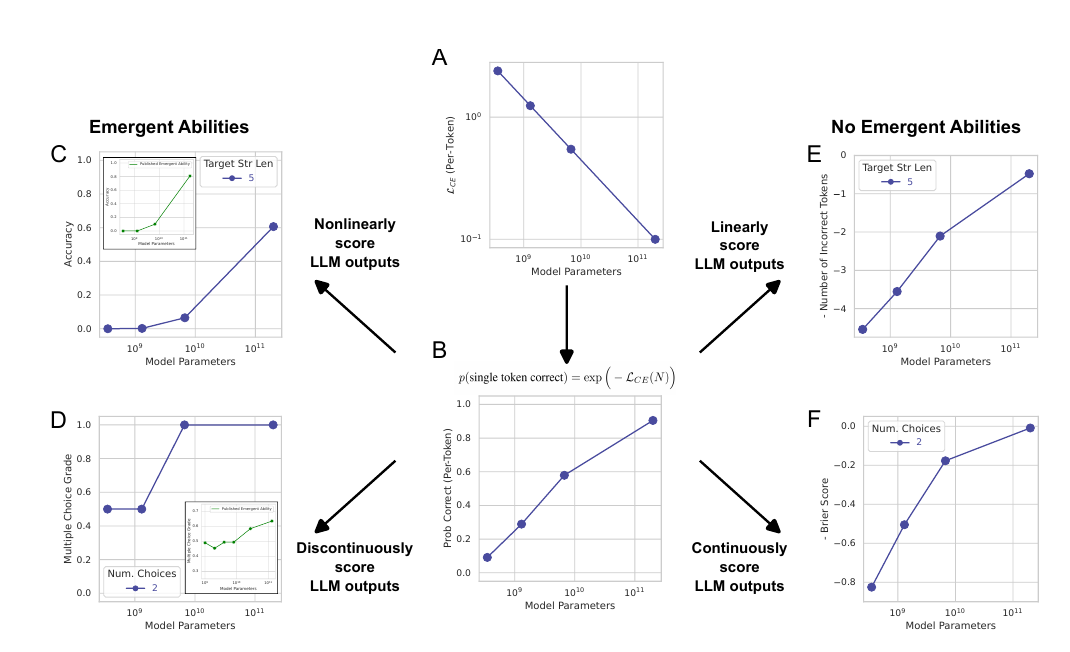}}
		\caption{Emergent abilities of LLMs are created by the chosen metrics \citep{schaeffer2023emergent}, not unpredictable changes in model behavior with scale.}
		\label{emg_abi}
	\end{center}
	\vskip -0.29in
\end{figure}
The appropriateness of employing various metrics to transform raw data on evaluation results remains uncertain. Nevertheless, more straightforward and dependable statistical methods can be employed for verification. From a statistical analysis standpoint, performance scores are considered as the dependent variable, with training parameters acting as independent variables. Other factors, such as training types and architecture, are regarded as random variables. This arrangement allows for the application of advanced regression models (e.g., GAMM) to conduct detailed testing and analysis.

 To elucidate the relationship among variables, we executed multiple sets of GAMM fittings. The first group aims to comprehend the impact of parameter count on performance scores while considering random effects. For this purpose, the specific GAMM equation employed is as follows:$ (log\_HellaSwag \sim s(log\_Param) + s(Architecture, bs =``re"), data = data)$. Here \texttt{s} is smooth, \texttt{re} = random effect.  The \texttt{s} in a GAMM model represents the smooth functions applied to predictors, allowing for flexible modeling of non-linear relationships in data, while also incorporating random effects to handle correlated groups or clusters within the data. In other words, the non-linear relationship in the data could be better detected using the smooth function. we applied a logarithmic transformation to the variables, which brought the data closer to a normal distribution. This transformation enables us to achieve more accurate fittings using GAMM. Furthermore, the addition of the random variable ``Type'' in the equation did not show significant results. This lack of significance could be attributed to the overshadowing effect of the ``Architecture'' factor, and ``Architectures'' is strongly significant across the evaluation datasets. The results are shown in Fig. \ref{gamm_plot}, where the parameter has a strongly significant effect on each of evaluation datasets (using a significance threshold defined by a \textit{p}-value less than 0.001). 
\begin{figure}
	\vskip 0.16in
	\begin{center}
		\centerline{\includegraphics[width=\columnwidth]{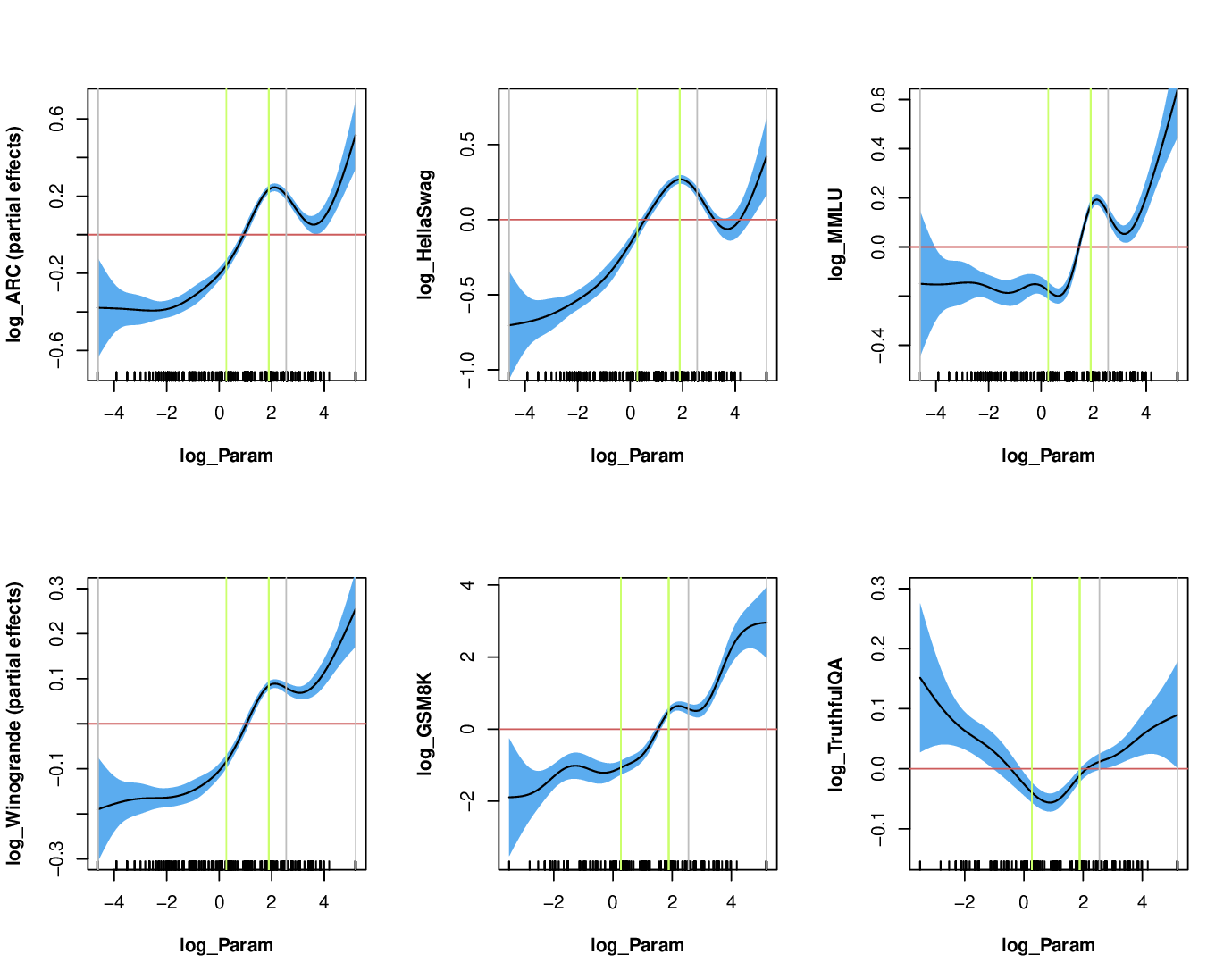}}
		\caption{Partial effects of parameters on LLMs performance scores. (In each plot, the vertical lines signify five quartiles, which divide the data on parameters into five equal quartiles: 0\%[0.01B] - 20\%[1.31B] - 40\%[6.53B] - 60\%[6.65B] -  80\%[12.85B] - 100\%[180B]. \textit{X-axis} ``log\_Param'' is the logarithm of model training parameters, and \textit{y-axis} is the logarithm of scores in each evaluation dataset. A curve of partial effects represents the relationship between a predictor variable and the response variable. Steeper slopes suggest a stronger relationship, and flatter slopes imply a weaker one. A curve of partial effects represents the relationship between a predictor variable and the response variable when a curve fluctuates around zero, it indicates that the effect is weak. The pointwise 95\%-confidence intervals are shown by the blue shadow.)}
		\label{gamm_plot}
	\end{center}
	\vskip -0.29in
\end{figure}

Referencing the standards illustrated in Fig. \ref{emg_abi} and the descriptions by \citet{wei2022emergent}, we observed discrepancies between the emergent abilities depicted and the findings in our Fig.\ref{gamm_plot}. Specifically, Panels E and F in Fig. \ref{emg_abi} lack flattened stages in their curves, which contradicts the principle of sharpness. Similarly, the curves in the six plots in Fig. \ref{gamm_plot} also show no flattened stages, aligning with the situation in Panels E and F of Fig. \ref{emg_abi}. However, while the abilities in Panels E and F of Fig.\ref{emg_abi} continually increase, the curves in Fig.\ref{gamm_plot} exhibit unpredictable changes between the 60\% and 100\% quartiles (6.65B-12.85B-180B). For ARC, HellaSwag, and MMLU, the curves exhibit a wave-like, non-linear fluctuation. In the first two cases, abilities at the final stage do not increase significantly post-fluctuation, leveling near the 60\%-80\% quartile range. MMLU shows no increase until the 20\% quartile. Winogrande and GMS8K, in contrast, display largely linear curves. TruthfulQA stands out with a unique curve pattern: a decrease at the initial stage, an increase between the 20\% and 80\% quartiles, yet a final level lower than the initial one, with unpredictability post-80\% quartile. Fig.\ref{gamm_plot} confirms that the parameter range \textbf{[3,7]}, as discussed in section 3.1 and analyzed using ANOVA and Tukey's method, demonstrates strong significance across multiple abilities. This parameter range consistently enhances scores across the evaluation datasets, as illustrated in Fig. \ref{gamm_plot}.
From Fig.\ref{gamm_plot}, the trends are partly consistent with ``emergent abilities'' descriptions. Notably, there is no initial stage flattening, implying the abilities do not emerge suddenly. However, from  the 60\% quartile, the four plots in Fig. \ref{gamm_plot} show some unpredictable tendencies, which could align with emergent abilities as described in the literature. It seems that ``GMS8K'' and ``Winogrande'' keep increasing with larger parameter scales. If ``emergent abilities'' imply an increasing trend with larger parameter sizes, then TruthfulQA's performance contradicts this, suggesting that such abilities might be present in certain parameter size ranges but not uniformly across all stages.

Additionally, due to the uneven proportion of some architectures (e.g., Llama-54\%, GPTNeo-11\%, GPT2-9\%, Bloom-3\% etc.), we scaled the data on the scores in different evaluation datasets in order to mitigate such unbalanced data. Based on the architecture categories, we calculated scaling factors to normalize the distribution across these categories. It then applied these scaling factors to various performance metrics in the dataframe. This kind of processing is useful in situations where models are fairly across different categories that might be unevenly represented in data. Using the same GAMM fittings, we can get the similar results as the data are not scaled. The visualization of the results is shown in Fig.\ref{para_fig1}, Fig.\ref{type_fig2}, and Fig. \ref{corr_fig2} in the \textbf{Appendix B}.

Our interest extends to exploring how different training types (such as fine-tune, instruction-tune, etc.) might influence the changes in abilities. To investigate this, we incorporated additional random effects into our analysis, yielding some notable observations. Specifically, we introduced the training type as a grouping factor in the GAMM fitting:  $(log\_HellaSwag\ \sim s(log\_Param, by=Type) + Type + s(Architecture, bs=``re"), data=data)$. ``$s(log\_Param, by=Type)$'' specifies a smooth function (\texttt{s}) of the logarithm of parameter, with a different smooth for each level of the ``Type'' variable. This allows for non-linear relationships between ``log\_Param'' and ``log\_HellaSwag'' that differ depending on the ``Type'', and it suggests that 'Type' is included as a categorical predictor in the model. The visualization is seen Fig. \ref{type_fig1} in the \textbf{Appendix B}. Our analysis revealed that the six evaluated abilities demonstrate significant results across both fine-tuning and instruction-tuning levels. However, in the context of RL-tuning, certain cases were identified as significant, characterized by a \textit{p}-value exceeding 0.05. In the scenario involving pretrained models, specifically for MMLU, the observed curve predominantly fluctuated around zero. These observations align coherently with the findings derived from ANOVA and Tukey tests, as detailed in Section 3.1.

The subsequent analysis delves into the interplay among various abilities. These models possess a range of capabilities, including language understanding, commonsense reasoning, and mathematical reasoning, which may interact and influence each other. To comprehend how a certain ability affects other abilities, we approached the analysis by considering the particular ability as the dependent variable, while treating the other capabilities as independent factors. This allows us to explore the potential impact some abilities may have on one given ability. To acheive this, we employed the following GAMM equation: $(log\_ARC \sim  s(log\_HellaSwag)+s(log\_MMLU)+s(log\_Winogrande)+s(log\_GMS8K)+s(log\_TruthfulQA) + s(Architecture, bs=``re"), data=data)$. For each ability, we employed a similar GAMM equation for estimation with random effects. When focusing on a particular ability as the dependent variable, that specific ability was excluded from the independent factors. The findings from this approach are presented in Fig.\ref{cor_eff} in the \textbf{Appendix B}. Our analysis revealed that a given ability influence other capabilities, while others do not certainly exert a significant impact (detailed in the \textbf{Appendix B}). To summarize, the abilities represented by ``HellaSwag'' and ``MMLU'' showed significant effects on other abilities, suggesting that knowledge reasoning and language understanding might play a pivotal role in influencing LLMs' overall capabilities. Specifically, both ``ARC'' and ``HellaSwag'' demonstrated a general influence across various abilities, as did ``MMLU''. In contrast, the remaining three abilities displayed insignificant effects (with \textit{p}-values greater than 0.05), indicating a lack of a general impact on the other capabilities. 

Additionally, the similar AVNOVA, Tukey and GAMM tests were done in the supplementary dataset, and the results are basically consistent with the ones in the primary dataset, The details are seen the \textbf{Appendix D}. 


\subsection{Clusters with key factors}

We employed t-SNE to compress data and facilitate clustering. The method's findings could check with the results from ANOVA and GAMM tests to some degree. For instance, certain parameter ranges did not form distinct clusters, and the clustering based on training types also lacked clear differentiation. The cluster results is shown in Fig.\ref{cluster_fig1} in the \textbf{Appendix D}. In Fig.\ref{cluster_fig1}, the first left panel indicates that the parameter range of [1.5,7] (in blue) forms a cluster. Conversely, the range of [0.01,1.5] forms a smaller cluster. Other parameter ranges, however, do not exhibit such clear clustering. The middle panel suggests that while fine-tuned models cluster to some extent, they are interspersed by instruction-tuned models. In the right panel, the Llama architecture appears to create a cluster.  Despite the formation of clusters, they are not distinctly separated from one another. This observation implies that clusters formed based on certain factors lack strong significance. It also suggests that the impact of specific parameter ranges, instruction-tuning, or certain architectures may not be as influential as anticipated. These clustering patterns, however, offer valuable cross-verification evidence that aligns with the results obtained from ANOVA and GAMM tests, contributing to a more comprehensive understanding of the data. 


\section{Discussion}

Our study presents findings that challenge certain established conclusions regarding the evaluation of LLMs in previous research. First, we question the purported superiority of instruction-tuning over fine-tuning. While previous studies (\citealp{wei2021finetuned}; \citealp{liang2022holistic}; \citealp{zhao2023survey}) indicate that instruction-tuned models generally outperform base models, our data does not support this assertion. According to our ANOVA and Tukey tests, no significant differences were observed between instruction-tune, fine-tune, and RL-tune across six evaluation datasets. Second, regarding the performance of small-sized, open-source models in mathematical reasoning, \citet{zhao2023survey} reported their underperformance. However, referring back to Fig. \ref{gamm_plot}, if we define the parameter range from the first to the third quartile as small-sized, these models exhibit comparable performance in mathematical reasoning tasks (e.g., GMS8K) to their larger-scaled counterparts. Third, \citet{zhao2023survey} claim that the 'Llama' model outperforms others is not corroborated by our analysis. We found that `Llama', along with other models like `GPT2', `Minstral', and `Falcon', show equivalent proficiency in complex reasoning tasks (GMS8K and ARC). Fourth, while \citet{zhao2023survey} suggested that scaling up open-source models consistently enhances performance, our findings indicate that this may be task-dependent. In the `TruthfulQA' task, increasing model parameters led to diminished performance. Moreover, as parameter sizes grow much larger, their effects become unpredictable. This suggests that scaling up models within a certain range can consistently improve performance, but beyond that range, the outcomes become uncertain.

Next, we discuss emergent abilities. The emergence of some advanced capabilities in LLMs might be attributed to their training, as inferred from comparisons with smaller-sized language models. However, the presence of certain abilities in the majority of LLMs does not necessarily imply that these abilities are intrinsic characteristics. As illustrated in Fig. \ref{gamm_plot}, these abilities manifest even with minimal parameters (as low as 0.01 billion).
Our research indicates a consistent increase in the capabilities of the models from the outset up to the point where 60\% of the data is utilized, aligning with the observations reported in \citet{schaeffer2023emergent}. However, the aspect of unpredictability, which our study identifies beyond the 60\% data threshold, was not observed or reported in the findings of \citet{schaeffer2023emergent}.  However, ``GMS8K'' and ``Winogrande'' seems to keep increasing, which are consistent with some research (\citealp{cobbe2021training}; \citealp{chowdhery2023palm}; \citealp{touvron2023llama}). It is possible that mathematical reasoning ability does not show unpredictable tendency when the parameter is up to 180B.
In the current study, a degree of unpredictability is evident when the parameter size exceeds 7B. In essence, the abilities of LLMs tend to scale almost linearly with parameter sizes up to 7B. Beyond this threshold, their performance becomes more predictable \citep{ganguli2022predictability}. The applicability of the scaling law appears to be confined within a specific range and may also vary depending on the nature of the tasks involved \citep{kaplan2020scaling}. 

While our findings diverge from previous research on emergent abilities in LLMs, this does not necessarily negate the concept entirely. Rather, it suggests a reevaluation: the abilities of LLMs continue to improve with increasing training sizes, yet performance becomes less predictable as the training size reaches extremely large scales. This indicates that simply expanding the training size may not be a consistently reliable method for enhancing the capabilities of LLMs \citep{patel2024scaling}.

In the following, we focus on the interplay among various abilities in LLMs. Our findings suggest that knowledge reasoning and language understanding could have an overall impact on the other capabilities of LLMs. \citet{burnell2023revealing} used the methods of factor analysis and correlation to identify three essential capabilities based on HELM \cite{liang2022holistic}: language comprehension, language modeling, and reasoning, and their finding is basically consistent with our findings. However, our study did not include the data on language modelling (e.g., word prediction, text generation). 

Finally, we discuss the issue concerning open-source LLMs and close-source ones. The \texttt{Open LLM Leaderboard} seems to feature evaluation outcomes from a limited number of close-source LLMs. Well-known close-source LLMs include GPT4, Gopher, Chinchilla, LaMDA, PanGu, Anthropic, and others. In contrast, open-source LLMs, exemplified by the over 120,000 models hosted on \texttt{Huggingface}, are significantly more abundant. The assessment of these close-source LLMs, especially those that are not freely available, on extensive evaluation datasets incurs considerable expense, presenting a substantial obstacle for researchers.
Despite this, the inclusion of extra data from a select few close-source LLMs in our analysis would likely not substantially impact the overall statistical outcomes. 
While some studies have focused on a handful of closed-source LLMs, often overlooking the majority of open-source counterparts, they assert that their findings from close-source models are representative of LLMs in general. This claim could be deemed contentious, given the larger proportion and accessibility of open-source LLMs.  

\section{Conclusion}
We applied a multifaceted statistical approach to investigate the influence of factors like scaling, training types, and architectures on the performance scores across different evaluation datasets for LLMs. Our analysis, however, did not yield conclusive evidence of a significant impact from these factors on enhancing LLM performance. Our research uncovered new characteristics of LLMs and shed light on the interactions between various abilities within these models. The statistical methods we adopted in this study not only provided clear and dependable ways to analyze real-world LLM performance data, but also hold potential for application in other domains within the realm of effective LLM evaluations.
\nocite{langley00}
\bigskip
\bigskip
\vspace{5in}
\bibliography{example_paper}
\bibliographystyle{apa}

\newpage
\appendix
\onecolumn
\section{ANOVA, Tukey Tests and Correlations}

The following details the significant parameter range scales on various evaluation datasets. Specifically, when evaluating the HellaSwag dataset, we observe that the differences between the [13,35] and [1.5,3] ranges, as well as between [7,13] and [3,7], are statistically insignificant. Similarly, in the ARC dataset, the difference between the [7,13) and [3,7) ranges is not significant. In the case of MMLU, the parameter ranges [1.5,3] and [0,1.5], [3,7] and [13,35], [7,13] and [3,7] are found to be insignificant. However, for TruthfulQA, the ranges [1.5,3] and [0,1.5] and [7,13] and [1.5,3] demonstrate significant differences, while other range comparisons do not yield significant results. In the Winogrande dataset, the comparisons between [3,7] and [13,35], [7,13] and [13,35], and [7,13] and [3,7] are not significant. Finally, in the GMS8K evaluation, the comparisons of [1.5,3) and [0,1.5], [3,7] and [13,35], [7,13] and [13,35], [7,13] and [3,7] do not exhibit significant differences. Intriguingly, the [3,7] range consistently shows significant differences when compared with certain other ranges across these datasets. This analysis underscores the nuanced relationship between parameter ranges and dataset performance, revealing critical insights into the varied impacts of these parameters across different benchmarks.

The following details the tests on architecture types. Table \ref{arch_sign} lists the pairs and their respective frequencies of significance using Tukey tests on score data on various architectures. 
Table \ref{arch_sign} shows that the frequent occurrence of these models in significant pairings across multiple benchmarks highlights their consistent and robust performance. This analysis demonstrates that certain architectures, specifically `GPT2', `Llama', `Bloom', `Mistral', and `GPTNeo', exhibit exceptional effectiveness in performing various evaluation tasks. 

\begin{table}
\begin{tabular}{lr}
	\toprule
	{} &  Frequency \\
	\midrule
	OPT-Llama      &          5 \\
	Llama-GPT2     &          5 \\
	Llama-Bloom    &          4 \\
	Mistral-GPT2   &          4 \\
	Other-Llama    &          4 \\
	Llama-GPTNeo   &          4 \\
	GPTNeo-GPT2    &          3 \\
	Other-GPT2     &          3 \\
	GPTJ-GPT2      &          3 \\
	Other-GPTNeo   &          3 \\
	Rwkv-Llama     &          3 \\
	Mistral-Bloom  &          2 \\
	Llama-Falcon   &          2 \\
	OPT-GPT2       &          2 \\
	OPT-Mistral    &          2 \\
	Llama-GLM      &          2 \\
	Mistral-GPTNeo &          2 \\
	Llama-GPTJ     &          2 \\
	Mistral-GPTJ   &          1 \\
	Rwkv-Mistral   &          1 \\
	Mistral-Llama  &          1 \\
	Mistral-GLM    &          1 \\
	Mistral-Falcon &          1 \\
	GPTNeo-GPTJ    &          1 \\
	OPT-GPTNeo     &          1 \\
	\bottomrule
\end{tabular}
\caption{The pairs and their respective frequencies of significance using Tukey tests on the score data considering the architecture types}
\label{arch_sign}
\end{table}

Table \ref{fig_cor_plot} illustrates the correlations among several abilities represented by their evaluation datasets. 

\begin{figure}
	\vskip 0.2in
	
		\centerline{\includegraphics[width=0.6\columnwidth]{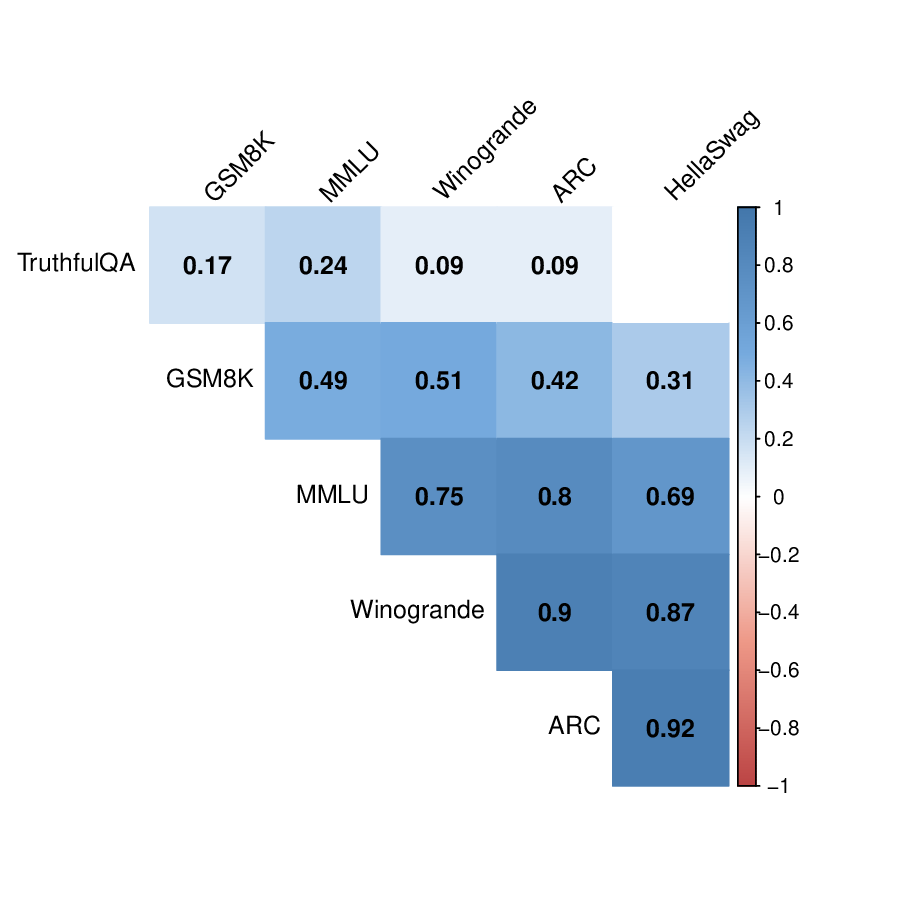}}
		\caption{Correlations among various benchmark datasets}
		\label{fig_cor_plot}
	\vskip -0.2in
\end{figure}

\section{GAMMs and the Partial Effects on Scaled Data}

First, let's delve into the workings of Generalized Additive Mixed Models (GAMMs). At their core, GAMMs operate based on the following fundamental mathematical equation:
\vspace{0.2cm}
$Y = \beta_0 + f_1(X_1) + f_2(X_2) + \ldots + f_n(X_n) + Zb + \epsilon$
\vspace{0.2cm}
\textit{Y} is the response variable.
\vspace{0.2cm}
$\beta_0$ represents the intercept.
\vspace{0.2cm}
$f_i(X_i)$ are the smooth functions of the predictor variables $X_1$, $X_2$, \ldots, $X_n$X 
\vspace{0.2cm}
\textit{Z} is the design matrix for the random effects.
\vspace{0.2cm}
\textit{b} symbolizes the random effects.
\vspace{0.2cm}
$\epsilon$ is the error term, typically assumed to be normally distributed.
Each $f_i$ is modeled using basis functions like splines, with complexity controlled by smoothing parameters. The random effects \textit{Zb} are assumed to follow a normal distribution, often with zero mean and a specific covariance structure.

This approach allows for a more intricate analysis of data relationships, as it combines multiple smooth functions with parametric elements. Thus, GAMMs are designed to discern a non-linear effect only when the data robustly suggests such a pattern. Conversely, it will identify a linear effect when the data predominantly supports a linear relationship.

\begin{figure}[ht]
	\vskip 0.2in
	\begin{center}
		\centerline{\includegraphics[width=0.89\textwidth]{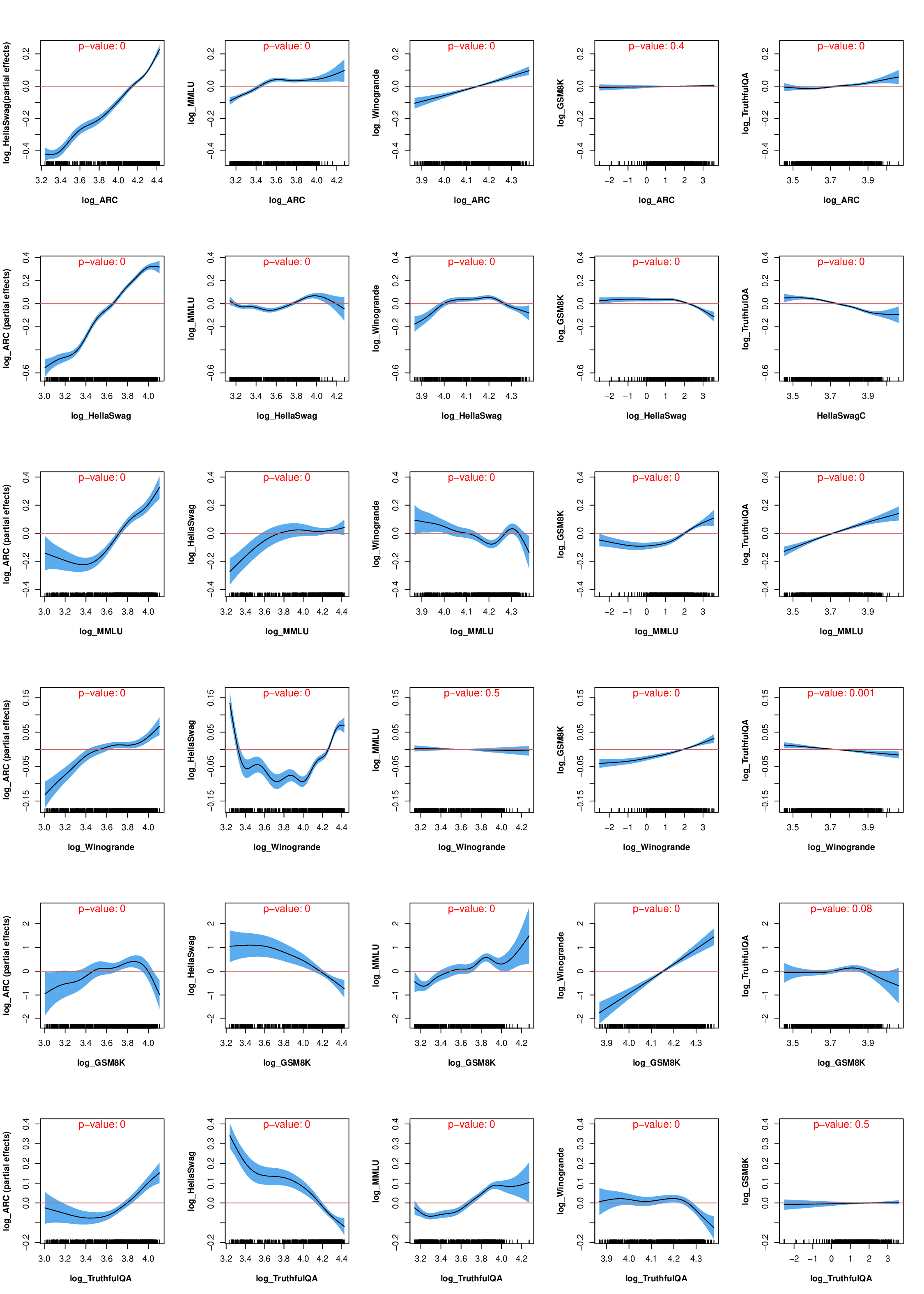}}
		\caption{Partial effects of one given ability on other abilities in LLMs. The \textit{x-axis} represents the logarithmic scale of a specific ability, while the \textit{y-axis} corresponds to the logarithmic scale of various other abilities. The slope of the curve provides insights into the strength of this relationship: a steeper slope indicates a more pronounced effect, whereas a gentler slope suggests a more subdued impact. Notably, when the value of \textit{p} is less than 0.01, the curve tends to level off near zero. This phenomenon signifies that the ability in question has little to no influence on the other ability.}
		\label{cor_eff}
	\end{center}
	\vskip -0.2in
\end{figure}

The following is the demonstration of visualization of GAMM analysis.  First, the visualization of one given ability on other abilities in LLMs is illustrated in Fig.\ref{cor_eff}.
Fig. \ref{type_fig1} shows how the parameter scale takes effect on the various abilities through different training types. As Fig. \ref{type_fig1} shown, several insignificant cases can be seen (i.e., \textit{p}-value is greater than 0.01). In our analysis, one particular instance within the RH-tuned model, when evaluated with ``TruthfulQA'', appears to be statistically insignificant. Additionally, the significance of the other two instances in the RH-tuned context is not particularly robust. This observation might suggest that the performance of the RH-tuned model has not meet the expectations. When examining the overall patterns exhibited in the fine-tune, instruction-tune, and pretrained models, we observe that their general curve shapes bear resemblance to each other. However, the specific abilities of these models demonstrate slight variations across the different training methodologies. Our findings from the GAMM analysis largely corroborate with the insights obtained from the ANOVA and Tukey tests, particularly concerning the distinctions among various training types. This consistency across different statistical methods reinforces the reliability of our results, highlighting nuanced differences in model performance based on the training approach. 

\begin{figure}
	\vskip 0.2in
	\begin{center}
		\centerline{\includegraphics[width=0.89\columnwidth]{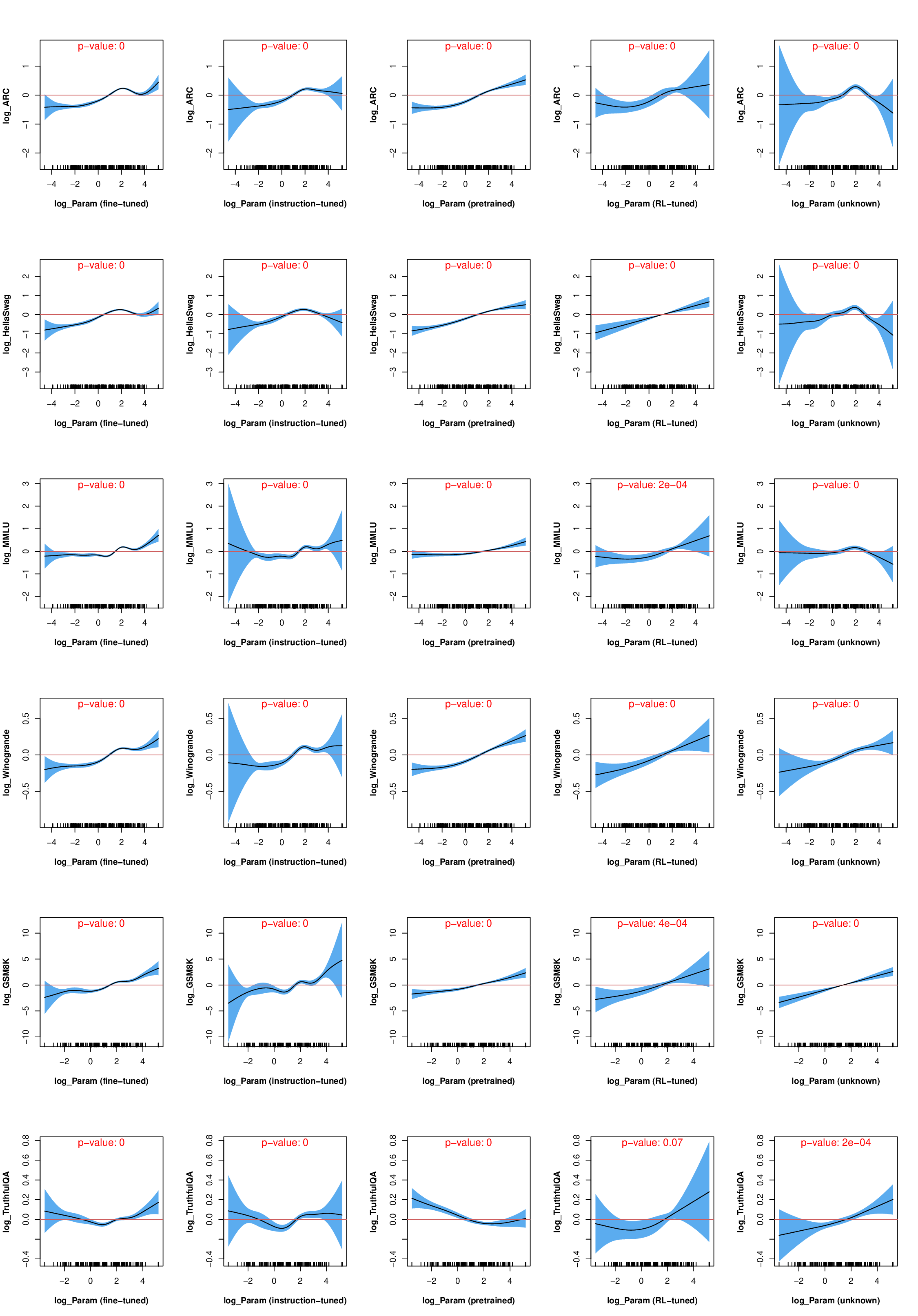}}
		\caption{Partial effects of parameters on LLMs performance scores across various training types}
		\label{type_fig1}
	\end{center}
	\vskip -0.2in
\end{figure}

Next, we visualized the GAMM results using the scaled data.
Fig. \ref{para_fig1} show how the parameter scale takes effect on the various abilities based on the scaled data which are normalized accordance to the frequency of architectures. Fig. \ref{para_fig1} are  consistent with Fig. \ref{gamm_plot}. 

We would like to delve deeper into the analysis of the ``TruthfulQA'' plot. The data shows a notable pattern based on the parameter range. Specifically, when the parameter is below the third quartile (i.e., 5.84B), there is a decrease in the score as the parameter size increases. In contrast, beyond the 5.84 B threshold, an improvement in performance is observed. Yet, this trend reverses with extremely large parameters (potentially exceeding 130B), where the score declines again. It is important to note that the peak scores, observed in the fourth and fifth quartiles, still fall short of the levels seen with smaller parameters. This pattern in ``TruthfulQA'' finds echoes in other research. For instance, Fig. 4 in \cite{lin2021truthfulqa} illustrates a similar trend in the 'Average Truthfulness' of generation tasks across five models, showing a decline in scores with larger model parameters. This phenomenon is also evident in the multiple-choice tasks of the same study. Moreover, \citet{zhao2023survey} observed in their ``TruthfulQA" evaluations that scores did not uniformly increase with larger model parameters. Such trends are not limited to ``TruthfulQA'' but seem prevalent in other Human Alignment tasks, such as C-Pair, WinoGender, RTP, HaluEVa. Based on these observations, it is hypothesized that other human alignment tasks might exhibit similar patterns to those demonstrated in ``TruthfulQA'' in our current analysis.


\begin{figure}
	\vskip 0.2in
		\centerline{\includegraphics[width=0.86\columnwidth]{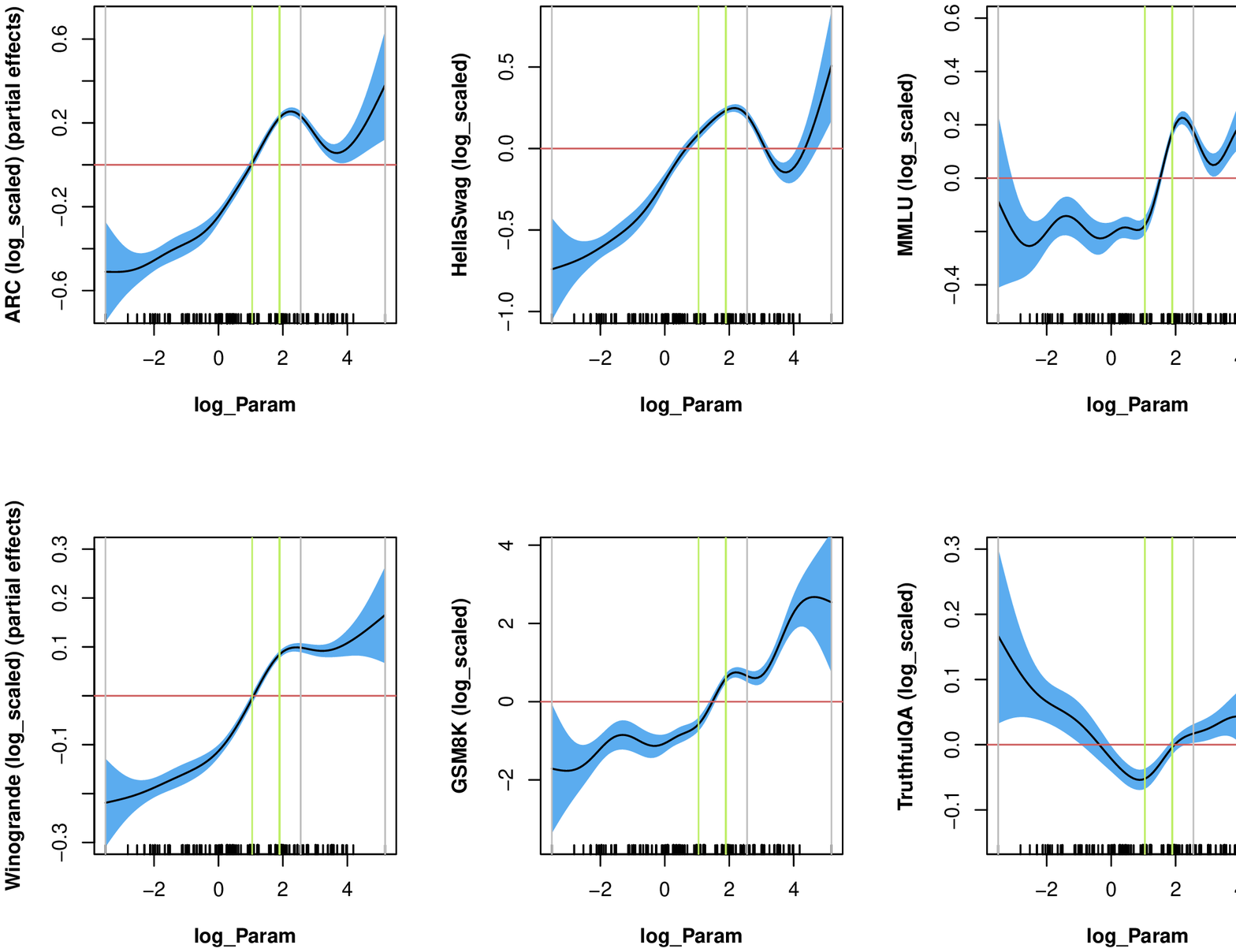}}
		\caption{Partial effects of parameters on LLMs performance scores on scaled data}
		\label{para_fig1}
	\vskip -0.26in
\end{figure}

Using the same scaling data, we plotted how the parameter scale takes effect on the various abilities through different training types, as shown in Fig. \ref{type_fig2}.  Fig. \ref{type_fig2} is basically consistent with Fig. \ref{type_fig1}.

\begin{figure}
	\vskip 0.2in
	\begin{center}
		\centerline{\includegraphics[width=0.89\columnwidth]{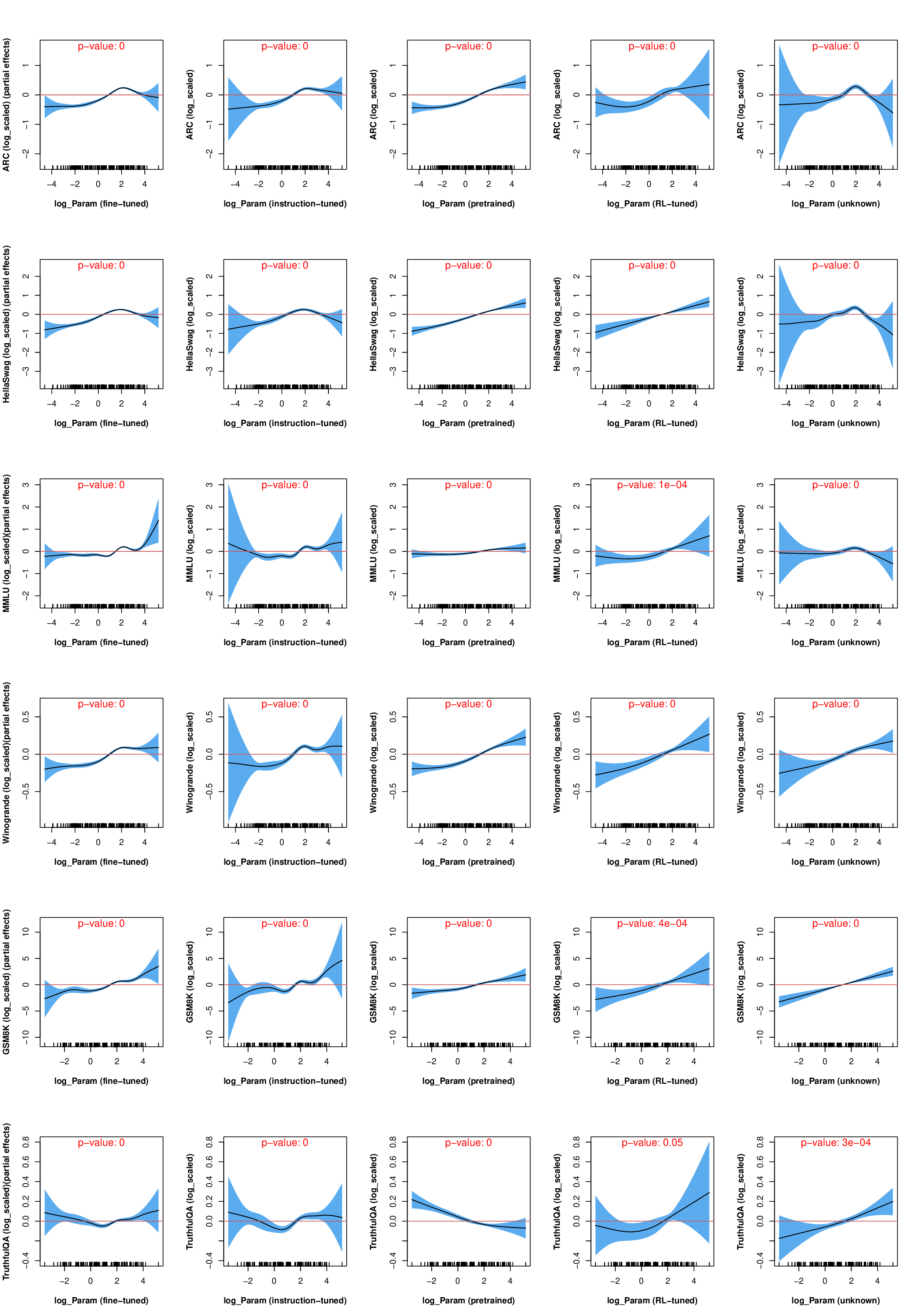}}
		\caption{Partial effects of one given ability on other abilities in LLMs on scaled data}
		\label{type_fig2}
	\end{center}
	\vskip -0.26in
\end{figure}

Meanwhile, using the same scaling data, we plotted how  the various abilities influence with each other, as shown in Fig. \ref{corr_fig2}.
\begin{figure}
	\vskip 0.2in
		\centerline{\includegraphics[width=0.86\columnwidth]{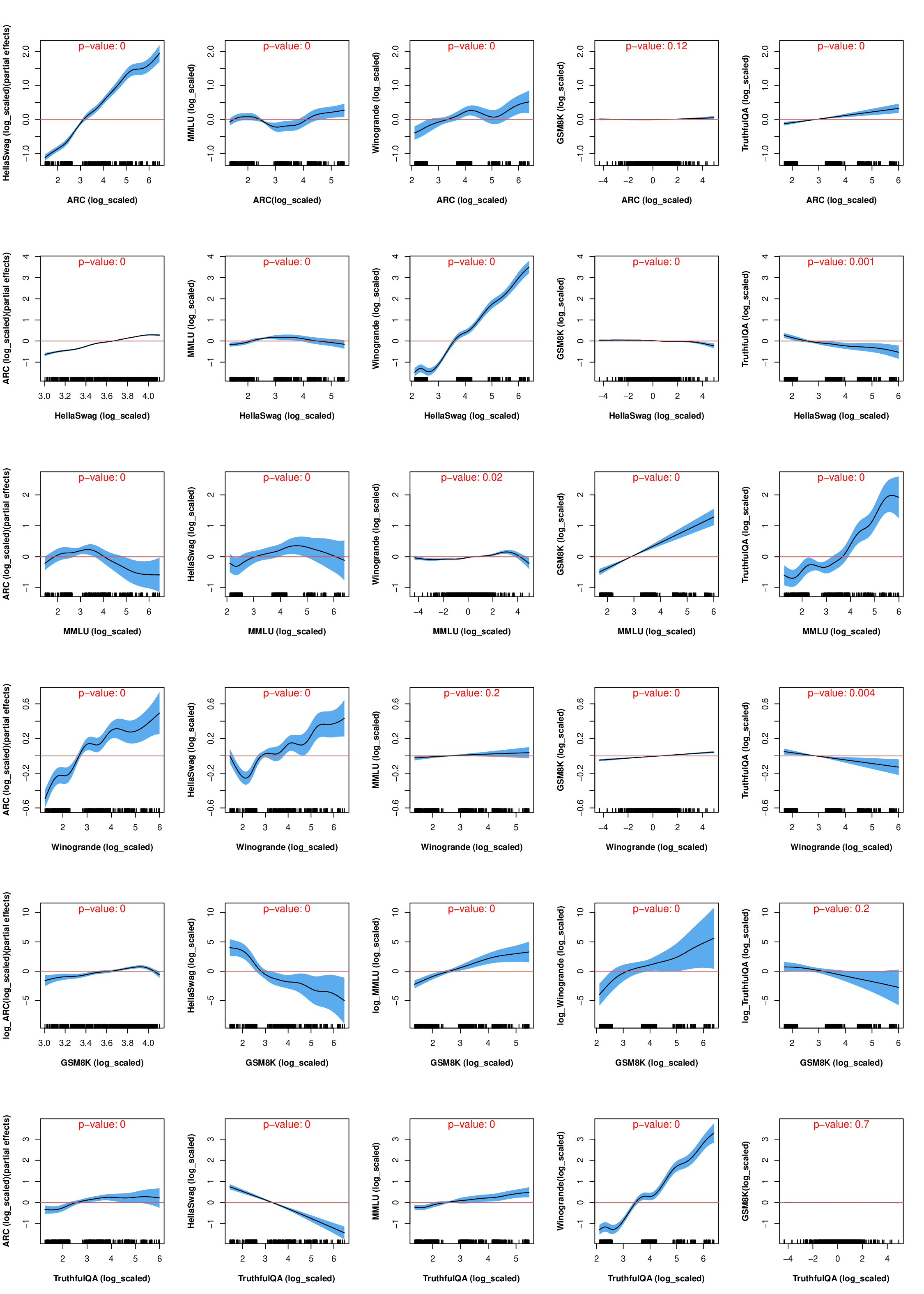}}
		\caption{Partial effects of parameters on LLMs performance scores across various training types using scaled data}
		\label{corr_fig2}
	\vskip -0.26in
\end{figure}

The T-sne clusters are shown in Fig. \ref{cluster_fig1}. The clusters of those claimed factors are not clearly separated, which are also consistent with the findings using Tukey and GAMM tests. 

\section{T-sne Clusters}
\begin{figure}
	\vskip 0.2in
	\begin{center}
		\centerline{\includegraphics[width=0.86\columnwidth]{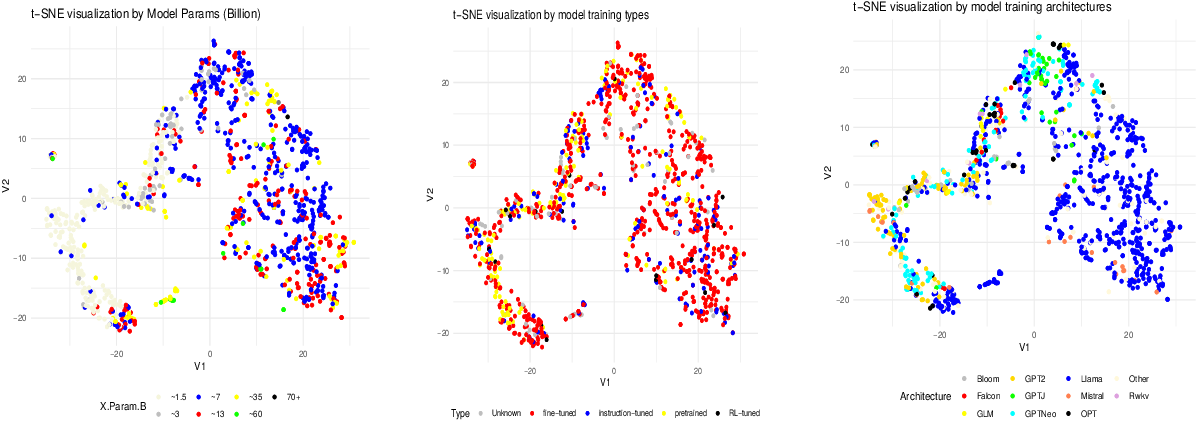}}
		\caption{T-sne clusters according to parameter range scale, training types, and architectures}
		\label{cluster_fig1}
	\end{center}
	\vskip -0.26in
\end{figure}

\section{Results of the supplementary dataset}

The supplementary dataset, which includes evaluation results of various LLMs, is accessible at \url{https://docs.google.com/spreadsheets/d/1kT4or6b0Fedd-W_jMwYpb63e1ZR3aePczz3zlbJW-Y4/edit?usp=sharing}. This dataset complements our primary dataset by providing additional information on training parameters and architectures. It encompasses models such as Flan, Pythia, GeoV, codegen, and ChatGPT, which were not included in our primary analysis. Furthermore, it features scores from five evaluation datasets: Lambada, Hellaswag, Winogrande, Piqa, and Coqa.

Each evaluation dataset is unique in its composition and testing methodology. For instance, `Lambada' contains 5153 passages sourced from books, where the model's task is to predict the final word of each passage based on its preceding context. `Piqa', on the other hand, comprises 1838 multiple-choice questions that test a model's commonsense physical intuition. The performance of the models on these datasets was measured using accuracy scores for Lambada, Hellaswag, Winogrande, Piqa, and F1 scores for Coqa. Further details can be found in the accompanying article \url{https://medium.com/@waiyan.nn18/understanding-and-benchmarking-evaluation-me\\trics-of-large-language-models-llms-in-2023-9a4a858f782b}.

Our statistical analysis involved applying ANOVA and Tukey's tests to examine the significance of data concerning parameter range scales and model architectures. This approach is consistent with how we processed the primary dataset. The architecture models in this study include 15 frameworks such as Bloom, Falcon, GPT-2, GPT-J, GPT-Neo, Llama, Mistral, OPT, Rwkv, ChatGPT, Flan, Pythia, GeoV, and codegen. The parameter ranges were segmented into five equal ranges: [0.7, 2.8], [2.8, 6.9], [6.9, 11], [11, 15.58], and [15.58, 135].

Significant findings were identified using a significance threshold defined by a p-value less than 0.05. Our analysis across various datasets indicated statistical significance varies across parameter ranges. Notably, the ranges [11, 15.58] and [15.58, 135] showed slight significance in `Coqa\_f1' scores. However, no significant results were found in other evaluation datasets through ANOVA tests. Additionally, models such as `Llama', `Pythia', `GPT-2', `GPT-J', `GPT-Neo', and `OPT' showed significant results in evaluating the datasets.

We applied a similar GAMM fitting to explore the effect of parameters on evaluation scores for each dataset: $(log\_Lambada \sim s(log\_Param) + s(Architecture, bs = ``re"), data = data)$. The results, visualized in Fig. \ref{core_fig11}, indicate that the effects of parameter range scales appear to be linear, without a flattened line at the initial stage, and without a predictable wave-shaped curve when the parameter size increases significantly. However, the curve trend in each plot is identical in Fig. \ref{core_fig11}, showing that the abilities keep increasing as the parameter becoming larger, and it is the same as ``emergent abilities'' described in \citet{schaeffer2023emergent}. However, these observations on continuously increasing abilities are merely based on the 65 LLMs in the supplementary dataset. We hypothesize that with a larger number of LLMs, the curve trend could replicate those observed in Fig. \ref{gamm_plot}.
 
\begin{figure}
	\vskip 0.2in
	\begin{center}
		\centerline{\includegraphics[width=0.86\columnwidth]{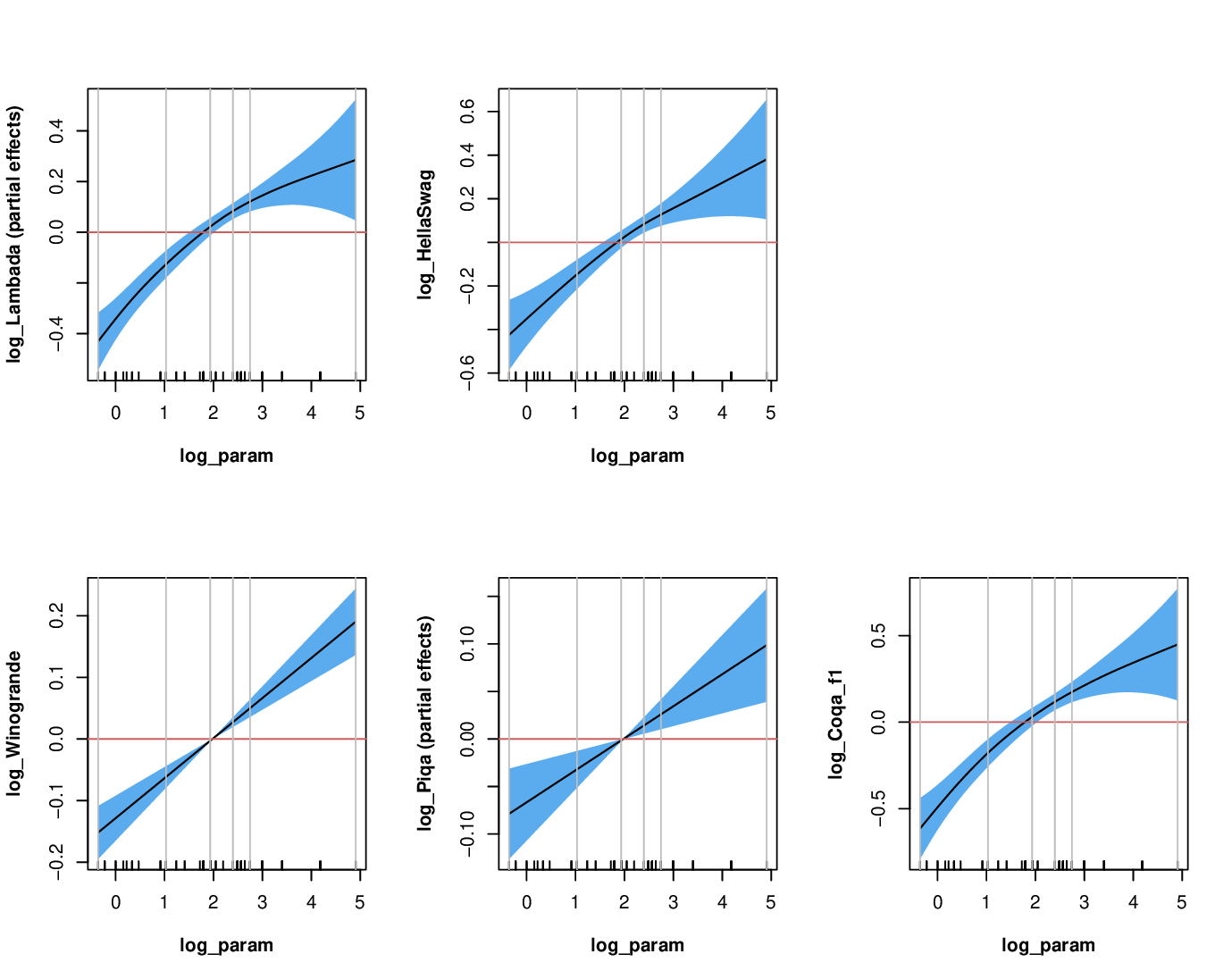}}
		\caption{Partial effects of parameters on LLMs performance scores from the supplementary dataset}
		\label{core_fig11}
	\end{center}
	\vskip -0.26in
\end{figure}

We also employed GAMMs to investigate the interplay between different abilities of language models. The GAMM framework was utilized to understand how one specific ability impacts others. The GAMM equation we applied is: $(log\_Lambada \sim  s(log\_HellaSwag)+s(log\_Winogrande)+s(log\_Piqa)+s(log\_Coqa) + s(Architecture, bs=``re"), data=data)$. For each evaluated ability, we formulated a similar GAMM equation with random effects for estimation. In cases where we focused on a specific ability as the dependent variable, that particular ability was omitted from the independent variables in the model. The outcomes of this analysis are depicted in Fig. \ref{corr_fig1}. Our findings reveal that the abilities tested in HellaSwag and Winogrande exhibit a notable influence on the performance in other abilities. This pattern of influence was not observed with the other abilities under consideration. This trend mirrors the observations from our primary dataset, where HellaSwag also demonstrated a general impact on the other abilities of the language models. Moreover, ``Lambada'' represents text generation has an overall impact on other abilities, which is consistent with the finding of language modeling (text generation) in \cite{burnell2023revealing}.
\begin{figure}
	\vskip 0.2in
	\begin{center}
		\centerline{\includegraphics[width=0.76\columnwidth]{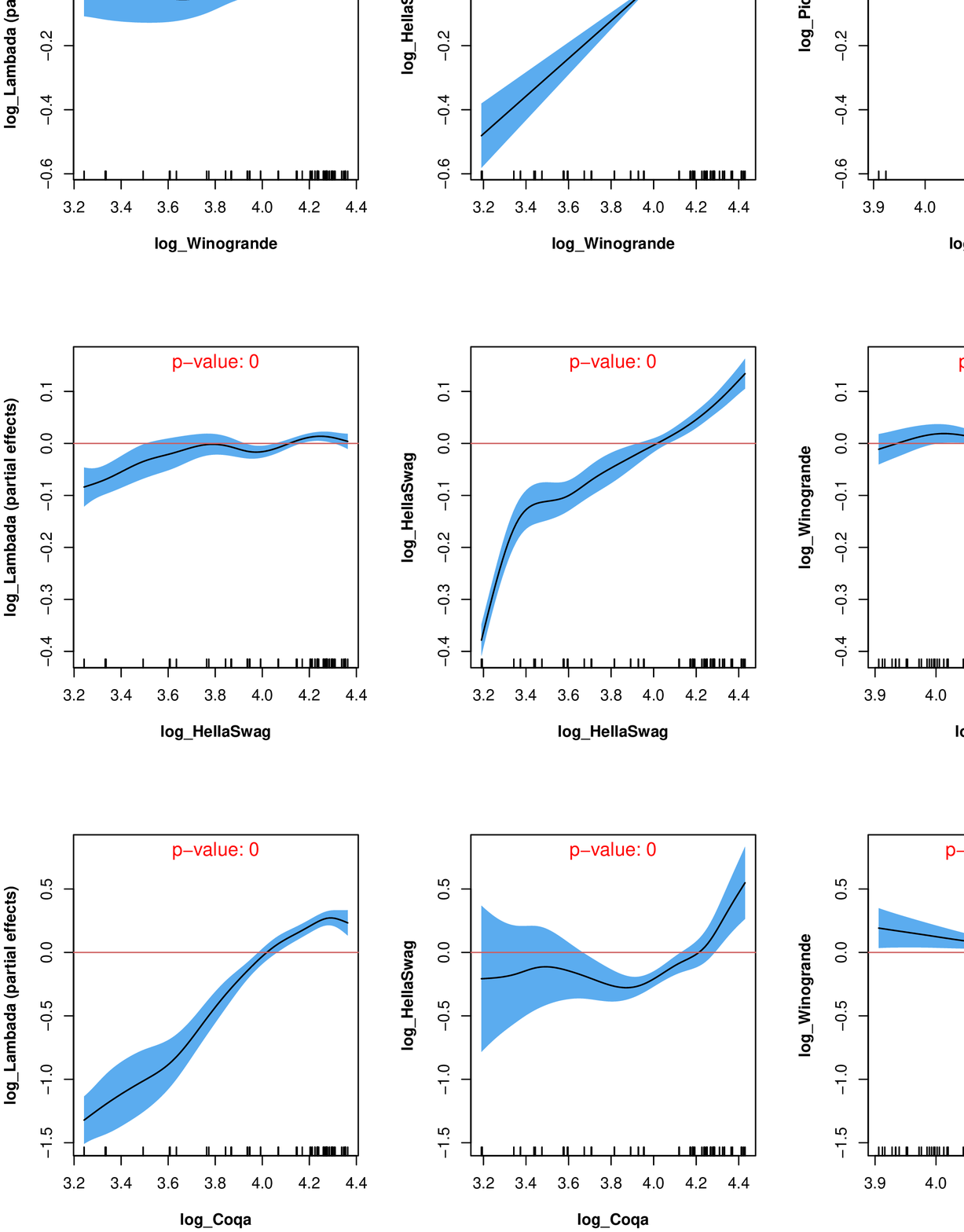}}
		\caption{Partial effects of one ability of LLMs on the other abilities in the supplementary dataset}
		\label{corr_fig1}
	\end{center}
	\vskip -0.26in
\end{figure}

\end{document}